\documentclass[conference]{ieeeconf}
\IEEEoverridecommandlockouts
\usepackage{cite}
\usepackage{amsmath,amssymb,amsfonts}
\usepackage{algorithmic}
\usepackage{graphicx}
\usepackage{textcomp}
\usepackage{xcolor}
\usepackage{multirow}
\usepackage{comment}
\usepackage{xcolor}
\usepackage{ulem}

\usepackage[absolute,overlay]{textpos}

\def\BibTeX{{\rm B\kern-.05em{\sc i\kern-.025em b}\kern-.08em
    T\kern-.1667em\lower.7ex\hbox{E}\kern-.125emX}}

\title{\LARGE \bf
Realtime Wind Estimation using Low Cost Quadrotor Uncrewed Aerial Vehicles
}

\author{Hiranya Udagedara$^{1}$ and Mahdis Bisheban$^{1}$
\thanks{This work was supported by the Natural Sciences and Engineering Research Council of Canada (NSERC), the Government of Alberta, Alberta Innovates, and the Schulich School of Engineering at the University of Calgary. Funding was awarded to Dr. Mahdis Bisheban, Director of the Intelligent Dynamics and Control Lab and Assistant Professor at the University of Calgary.}
\thanks{$^{1}$The authors are with the Department of Mechanical and Manufacturing Engineering, University of Calgary, Calgary, Canada
        {\tt\small hiranya.udagedara@ucalgary.ca, mahdis.bisheban@ucalgary.ca}}}

\begin{document}

\begin{textblock*}{\textwidth}(2.0cm,0.5cm)
\centering
\small Author's preprint. Published by IEEE American Control Conference (ACC), 2026.

\end{textblock*}

\maketitle

\begin{abstract}
In environmental monitoring as well as emergency response applications such as wildfires, wind velocity measurement is essential. Quadrotor UAVs have become popular platforms for wind velocity estimation due to their maneuverability, compact size, and cost-effectiveness. Numerous studies use the Extended Kalman Filter (EKF) to estimate the wind velocity based on the quadrotor dynamic model. However, most of them use hovering quadrotors only for wind estimation, others use a near-linear trajectory to estimate near-constant velocities. Furthermore, EKF performance is constrained by its reliance on linearized approximations of the nonlinear quadrotor dynamics around current states, limiting accuracy in highly nonlinear scenarios, including windy conditions. This study proposes the use of an Unscented Kalman Filter (UKF), a nonlinear estimator to provide accurate wind estimations while maintaining the trajectory of the quadrotor UAV. The quadrotor is modeled on the Special Euclidean group \(SE(3)\) and the approach is evaluated through numerical simulations using a geometric controller to maintain quadrotor flight paths.
The results indicate that as the nonlinearity of the simulation increases, the UKF consistently outperforms the EKF. This demonstrates the potential of the UKF as a reliable estimator for highly nonlinear scenarios, capable of maintaining the trajectory with minimal deviation while providing accurate wind velocity estimations.
\end{abstract}

\section{Introduction}
Real-time wind velocity estimation is very important in environmental data collection. In addition, in extreme cases, such as the monitoring of wildfires, knowing the wind velocity allows more buffer time for preparation. Although wind collection centers are ideal for long-term data collection, for emergencies such as wildfires, portability is important. Quadrotor UAVs are essential in this context. Quadrotor UAVs are better than fixed-wing UAVs as they are more maneuverable, cost-effective, and compact in size.

Direct measurement and model-based estimation are the two main methods for estimating wind velocity. Direct measurement requires additional sensors, such as anemometers, which significantly increase the cost of low-cost drones. These sensors also add weight, thereby reducing flight endurance, and require information about its position and velocity using onboard measurements. In contrast, model-based estimation avoids the need for extra hardware by inferring wind velocity from the drone’s dynamics and onboard measurements, making it more cost-effective and lightweight.

Some researchers have used the Kalman filter to estimate wind velocity \cite{Neumann2015, Xiang2016}. However, the Kalman filter assumes a linear quadrotor UAV model. Therefore, its capabilities limit maneuverability and accuracy of trajectory tracking and precision. In contrast, the EKF approximates the nonlinear quadrotor UAV system by linearizing it around the current states. Gonzalez et al. have used EKF to estimate the 2D wind velocity using real-time data, where the data are post-processed \cite{Gonzlez-Rocha2019}. Sikkel et al. estimated the wind velocity in real-time with a hovering quadrotor UAV, under near constant wind disturbances \cite{LNCSikkel2016}. In the work done by Xing et al., wind velocity estimation was carried out in a simulation environment, with a quadrotor UAV hovering in a wind field created by a Dryden model \cite{Xing2017}. The work has taken a different approach, as they have estimated the wind velocity by parameterization of an equation for wind velocity. The EKF has been used to approximate the parameters \cite{Xing2017}. The work by Chen et al. uses an invariant EKF for 3D wind velocity estimation. The vertical wind velocity is estimated using a hovering quadrotor UAV \cite{Chen2022}. UKF is another variant of the Kalman filter, which is a nonlinear estimator. In the work by Condomines et al., the UKF has been used with a fixed-wing UAV for wind velocity estimation, which estimates the wind by direct measurement \cite{Condomines2015, Rhudy2017}. Shastry et al. have used a variant of UKF, sq-UKF, for 3D wind velocity estimation \cite{Shastry2021}. They have estimated wind velocity in a simulated environment and estimated wind gust velocity while the quadrotor is in a hovering position. In a later work, they have used a wind turbine to simulate wind, which varies in one direction \cite{Shastry2023}.

The use of EKF for wind velocity measurements with quadrotor UAVs is widely discussed. However, the use of UKF for wind velocity estimation has yet to be explored.
 With recent advances in low-cost onboard computing, the previously prohibitive computational demands of the UKF no longer pose a significant limitation.
However, in both EKF- and UKF-based studies focus has been mainly on the planar wind velocity \cite{Xing2017, Gonzlez-Rocha2019, LNCSikkel2016}. Only a few studies focus on all three components of wind velocity \cite{Chen2022,Shastry2021,Shastry2023}. Among them, only Chen et al. performed real experiments and using a quadrotor UAV for wind estimation \cite{Chen2022}. But the flight is a straight-line trajectory, and the vertical wind velocity is non-existent. Therefore, even with EKF, wind velocity estimation while performing complex trajectories is to be explored.

In this study, the main focus is on precise and accurate wind velocity estimation during trajectory tracking, using low cost quadrotor UAVs, while using the same algorithm for state estimation, without requiring extra computation and power recourse. This enables the quadrotor UAV to monitor the wind velocity while continuing on its intended flight. This allows for a more stable and energy-efficient flight while collecting environmental data. A model-based approach is used to estimate the wind velocity. 
Measurements are obtained from commonly available onboard sensors, GPS, and a 6-axis IMU (3-axis gyroscope and 3-axis accelerometer).

A wide range of models have can be used for wind estimation. These models span from simple static relationships, such as those linking power or thrust and the UAV’s orientation or tilt to wind speed \cite{Xiang2016}, to more sophisticated approaches, including kinematic and/or dynamic particle models \cite{Neumann2015}, translational kinematic and dynamic models of a rigid body \cite{Gamagedara2019}, and orientation-based kinematic and dynamic models\cite{Shastry2021}. Orientation-based kinematic and dynamic models mostly represented by minimal representation like Euler angles \cite{Rhudy2014}. In general, dynamic models are required for estimating time-varying wind or when the UAV is in motion. By contrast, in the case of a hovering UAV under constant wind conditions, static or particle kinematic and dynamic models can provide estimation up to some wind magnitude \cite{gonzalez2019sensing}. In this paper, we use both translational and orientation kinematic and dynamic model of a quadrotor UAV on SE(3) to avoid singularity problem. To the best of author knowledge this is the first time, one uses these complete models for realtime 3D wind estimation.

In this paper, Section II presents the methodology, followed by the results and discussion in Section III. The paper concludes with Section IV.

\section{Methodology}
This section presents the quadrotor UAV's dynamic model, the UKF and EKF estimation processes, and experimental details.

\subsection{Dynamic Model of the Quadrotor UAV}
This section presents the dynamic model of the quadrotor UAV on SE(3). The origin of the body frame is assumed to be located at the center of mass of the quadrotor. Without loss of generality, the center of mass of the IMU is assumed to coincide with that of the quadrotor. In the case of any misalignment in orientation or displacement between the IMU and the center of mass of the UAV, these can be corrected through calibration of the IMU. As the body frame changes its orientation, the IMU frame remains fixed with respect to the body frame. Let \(x\in \mathbb{R}^3\) represent the position of the origin of the body frame in the inertial frame. Then \( v \in \mathbb{R}^3 \) represents the velocity of the origin of the body frame in the inertial frame. The angular velocity of the quadrotor in the body frame is represented by \( \Omega \in \mathbb{R}^3 \), and its orientation is denoted by \( R \in \mathbb{R}^{3 \times 3} \).  \(v_w \in \mathbb{R}^3\) is the wind velocity in the inertial frame. In this study, 21 parameters are considered, thus, the full state vector \( X = [x,v,\Omega,R, v_w]\in \mathbb{R}^{21} \).
The translational and attitude dynamics of a quadcopter subject to a wind disturbance are as follows:
\begin{align}
    \dot{x} =& v,
    \label{eq:xdot_w1}\\
        \dot{v}_b =& v_b \times \Omega + R^Tge_3 - \frac{f_ce_3}{m} + R^T\dot{v}_w + \frac{D_f}{m} + R^T\frac{\Delta_1}{m},\label{eq:vbdot_w1}\\
    \dot{R} = & R\hat{\Omega},
    \label{eq:Rdot_w1}\\
    \dot{\Omega} =& J^{-1}(-\Omega \times J\Omega + M_{c} + \Delta_2),
    \label{eq:Wdot_w1}
\end{align}
where \(e_3=[0, 0, 1]^T \in \mathbb{R}^3\), and \(v_b \in \mathbb{R}^3\) is the vehicle velocity in the body frame.  The velocity \(v_b\) can be expressed in terms of the inertial velocity \(v\) as, \(v_b = R^Tv\). Thus, \(\dot{v}\) can be defined as follows,
\begin{align}
    \dot{v} =& ge_3 - \frac{f_cRe_3}{m} + \dot{v}_w + \frac{RD_f}{m} + \frac{\Delta_1}{m}.\label{eq:vdot_w1}
\end{align}
In this study, the velocity of the quadrotor in inertial frame \(v\) is not measured directly, but estimated using measurements from the GPS. Thus, Equations \eqref{eq:xdot_w1} to \eqref{eq:vdot_w1} are used.
A geometric controller is employed to generate the thrust force (\(f_c \in \mathbb{R}\)) and the moment vector (\(M_c \in \mathbb{R}^{3}\))  necessary to maintain the quadrotor on its designated path \cite{Goodarzi2017}, which are assumed to be instantly realized. The rate of change of wind velocity was assumed to remain constant until the next time step \((\dot{v}_w = [0, 0,0]^T)\), which is acceptable for small time-steps. The symbol  \(\hat{(.)}\)  represents the hat map.  
\(\Delta_1 \in \mathbb{R}^3\) and \(\Delta_2 \in \mathbb{R}^3\) account for the external forces and torques in the body frame, which are not explicitly modeled. 
\(\Delta_1\) and \(\Delta_2\) are considered fixed but unstructured. 
  
In Equation \eqref{eq:vdot_w1}, \(D_f \in \mathbb{R}^3\) denotes the drag force, and the generic horizontal drag force is given by
\begin{align}
    D_f = -\frac{1}{2}\rho_{air}C_d|v_r|v_r,
    \label{eq:dragf_w1}
\end{align}
where \(C_d \in \mathbb{R}^{3 \times 3}\) is the body drag coefficient matrix. The relative wind velocity in the body frame \(v_r \in \mathbb{R}^3\) is given by \(v_r = v_b - R^Tv_w\).

\subsection{Measurement}
It is assumed that \(x_{gps}\), \(\Omega_{imu}\), and \(a_{imu}\in \mathbb{R}^ 3\) are obtained from the low cost sensors, including the GPS and the IMU. The measurement vector is \(Y=[x_{gps}, \Omega _{imu}]\in \mathbb{R}^6\), which are available on low cost UAVs. Acceleration measurement \(a_{imu}\) is directly used in updating the wind velocity \(v_w\).

\subsection{Estimation}
Estimation of the full state vector \(X\) is performed using both EKF and UKF. Both estimators carry out the estimation in two sections: flow update and measurement update. During flow update, the full state vector \(X_{k-1}\) and error covariance matrix \(P_{k-1}\) of the previous time step are propagated. The propagated full-state vector \(X_k^-\) and the error covariance matrix \(P_k^-\) are then adjusted during the measurement update. The subscript \({k-1}\) refers to the previous time step. The superscript \(^-\) indicates the prior estimate, and \(^+\) indicates the posterior estimate. 

In the literature, wind dynamics are often modeled as a stochastic process, for example using Random Walk or Gauss-Markov \cite{Rhudy2019}. Alternatively, many studies assume \(\dot{v}_w=[0,0,0]^T\), where the wind velocity is directly measured. In this study, a different approach is followed, as the wind dynamics are modeled under the assumption \(\dot{v}_w\) is negligible and no stochastic model is introduced. In contrast, at each time-step, the priori estimate of the wind velocity \(v_{w_k}^-\) is obtained using the relationship in Equation \eqref{eq:dragf_w1}. The drag force \(D_f\) is derived using the relationship in Equation \eqref{eq:vdot_w1}. The reading from the accelerometer \(a_{imu}\) is such that,
\begin{align}
    a_{imu} = \dot{v}_b - R^Tge_3 = v_b \times \Omega - \frac{f_ce_3}{m} + \frac{D_f}{m} + R^T\frac{\Delta_1}{m}.
\end{align}
In this formulation, the accelerometer is assumed to measure the non-gravitational acceleration (specific force) in the body frame, i.e., the acceleration relative to the world frame with gravity neglected. In an ideal situation, where there is no wind disturbance, the acceleration measured by the accelerometer would be as follows,
\begin{align}
    a_{ideal} = v_b \times \Omega - \frac{f_ce_3}{m} + R^T\frac{\Delta_1}{m}.
\end{align}
Thus, the drag force is roughly, the difference between the two accelerations, \(a_{imu}\) and \(a_{ideal}\). Therefore, at each time-step, the measurement from the accelerometer \(a_{imu}\) is used to calculate the supposed drag force as follows,
\begin{gather}
    D_{f_k} = m(a_{IMU_k}-a_{ideal_k}),
    \label{eq:Dragf_WI1}
\end{gather}
 By calculating \(D_{f_k}\), the wind velocity \(v_{w_k}\) can be estimated using \eqref{eq:dragf_w1}. This prior estimate of the wind velocity \(v^-_{w_{k}}\), is then refined during measurement update step.

\paragraph{EKF}
The EKF algorithm relies on a state-dependent linearization of the nonlinear quadrotor system given in Equation \eqref{eq:xdot_w1} to \eqref{eq:Wdot_w1}. For that Jacobian-based linearization is carried out by introducing perturbed states \(\delta \dot{X} = [\delta \dot{x}, \delta \dot{v}, \delta \dot{\Omega}, \delta \dot{R}, \delta \dot{v}_w]\). In geometric control, the rotation matrix \(R\) is not directly perturbed, as it should remain on \(SO(3)\). Instead, we express perturbations with a minimal vector \(\eta \in \mathbb{R}^3\) through the exponential map \cite{Gamagedara2019}.
Using first-order approximation, \(\eta\) is defined so that it satisfies,
\begin{equation}
       \delta R = \left. \frac{d}{d\epsilon} \right|_{\epsilon=0} R \exp(\epsilon \hat{\eta}) = R \hat{\eta},
   \label{eq:dR}
\end{equation}
where \(\epsilon \in \mathbb{R} \) is a smaller scalar parameter controlling perturbation magnitude. Therefore, the perturbed state vector \(\delta \dot{X} = [\delta \dot{x}, \delta \dot{v}, \delta \dot{\Omega}, \delta \dot{\eta}, \delta \dot{v_w}]\in \mathbb{R}^{15}\) is used in obtaining the Jacobian linearization.  The system was linearized around \(v_w = 0\), as the nominal wind is assumed unknown and also to simplify the analytic Jacobians. Thus, perturbed form of Equation \eqref{eq:xdot_w1} can be expressed as follows,
\begin{align}
    \delta \dot{x} &= \delta v. \label{eq:dxdot_w1}
    \end{align}
    The perturbed form of Equation \eqref{eq:vdot_w1} can be expressed as follows,
    \begin{align}
     \delta \dot{v} &= \delta \big(- \frac{f_cRe_3}{m}\big) +  \delta \big(\frac{RD_f}{m}\big),\notag
     \end{align}
where, \(RD_f = -\frac{1}{2}\rho_{air}C_dR|R^Tv-R^Tv_w|(R^Tv-R^Tv_w) = -\frac{1}{2}\rho_{air}C_d|v-v_w|(v-v_w)\). Thus, \(\delta \dot{v}\) can be expressed as,
\begin{align}
    \delta \dot{v} &=   -\frac{1}{2m}\rho_{air}C_d(|v|I_{3\times 3} + \frac{vv^T}{|v|})\delta v + \frac{f_c}{m} R(e_3)^{\wedge}\eta,  \label{eq:dvdot_w1}
    \end{align}
    where \(|v|=\frac{v^T}{|v|}\), and \( \delta (Re_3)= R\hat{\eta}e_3 =  R(e_3)^{\wedge}\eta\). As \(\eta\) is already a perturbation itself, \( \dot{\eta} = \delta \dot{\eta}\). Therefore, using Equation \eqref{eq:Rdot_w1} and \eqref{eq:dR}, \(\dot{\eta}\) can be defined. From Equation \eqref{eq:Rdot_w1}, 
    \begin{align}
        \delta \dot{R} = R\hat{\eta}\hat{\Omega} + R(\delta \hat{\Omega}). \notag
    \end{align}
    From the relationship in Equation \eqref{eq:dR},
    \begin{align}
        \delta \dot{R} &= \dot{R}\hat{\eta}+R\hat{\dot{\eta}}. \notag
    \end{align}
    By equating both expressions for \(\delta \dot{R}\), and substituting from Equation \eqref{eq:Rdot_w1}
    \begin{align}
        \dot{R}\hat{\eta}+R\hat{\dot{\eta}} &= R\hat{\eta}\hat{\Omega} + R(\delta \hat{\Omega}), \notag \\
        R\hat{\dot{\eta}} &=  R(\delta \hat{\Omega}) + R\hat{\eta}\hat{\Omega} - R\hat{\Omega}\hat{\eta}, \notag \\
        \hat{\dot{\eta}} &= (\delta \hat{\Omega}) + \hat{\eta}\hat{\Omega} - \hat{\Omega}\hat{\eta}. \notag
    \end{align}
    Given that \(-\hat{\eta}\hat{\Omega} + \hat{\Omega}\hat{\eta} = \widehat{\Omega \times \eta}\) ,
\begin{align}
    \hat{\dot{\eta}} & = \delta \hat{\Omega} - \widehat{\Omega \times \eta}, \notag \\
    {\dot{\eta}} & = \delta {\Omega} - {\Omega \times \eta}, \notag \\
    \dot{\eta} &= -\hat{\Omega}\eta + \delta \Omega,  \label{eq:detadot_w1}
    \end{align}
where, \(\Omega \times \eta = \hat{\Omega}\eta\). The perturbed state \(\delta \dot{\Omega}\) is obtained by linearizing the rotational dynamics given in  Equation \eqref{eq:Wdot_w1} and neglecting the first-order variations of the control/disturbance torque yields, as follows, 
    \begin{align}
    \delta \dot{\Omega} &= \delta  (J^{-1}(-\Omega \times J\Omega + M_{c} + \Delta_2)),\notag\\
    \delta \dot{\Omega} &= J^{-1}\delta (-\Omega\times J\Omega). \notag 
    \end{align}
    As \(-\Omega\times J\Omega = -\hat{\Omega}J\Omega = (J\Omega)^\wedge\Omega\),
    \begin{align}
    \delta \dot{\Omega} &= J^{-1}(-\hat{\Omega}J\delta\Omega + (J\Omega)^\wedge\delta\Omega), \notag\\
        \delta \dot{\Omega} &= J^{-1}((J\Omega)^\wedge - \hat{\Omega}J)\delta \Omega.  \label{eq:dwdot_w1}
\end{align}
    ***  Look into \(f_c\) and \(M_c\) by referring Dr. Bisheban's thesis ***
    The perturbed wind dynamics can be expressed as follows,
    \begin{align}
    \delta \dot{v}_w &= 0.  \label{eq:dvwdot_w1}
\end{align}
 The second-order perturbations have been neglected in this study. At the current time-step, Equations \eqref{eq:dxdot_w1} to \eqref{eq:dvwdot_w1}, the quadrotor dynamics can be expressed in the linearized form \(\delta \dot{X}=F_{t_{k-1}}\delta X+G_{t_{k-1}}\mathcal{W}\), where the state transition matrix \(F_{t_{k-1}} \in \mathbb{R}^{15 \times 15}\) and process noise transition matrix \(G_{t_{k-1}}\in \mathbb{R}^{15 \times 6}\) can be expressed as follows, 
 \begin{align}
        F_{t_{k-1}} =
    \begin{bmatrix}
        0_{3} & I_{3} & 0_{3} & 0_{3} & 0_{3} 
        \\
        0_{3} & F_{1} & F_{2} & 0_{3} & 0_{3}  
\\
        0_{3} & 0_{3} & F_{3} & I_{3} & 0_{3}   
\\
        0_{3} & 0_{3} & 0_{3} & F_{4} & 0_{3} 
\\
        0_{3} & 0_{3} & 0_{3} & 0_{3} & 0_{3} 
\\
    \end{bmatrix},
    G_{t_{k-1}} = 
    \begin{bmatrix}
        0_{3} & 0_{3}  \\
        \frac{I_{3}}{m} & 0_{3} \\
        0_{3} & 0_{3}  \\
        0_{3} & J^{-1} \\
         0_{3} & 0_{3}
    \end{bmatrix}, \label{eq:G_WI1}
\end{align}
where,
\begin{align*}
    F_{1} =&   -\frac{1}{2m}\rho_{air}C_d(|v_{k-1}|I_{3\times 3} + \frac{v_{k-1}v_{k-1}^T}{|v_{{k-1}|}}),\\
    F_{2} =&  (f_{c_{k-1}}/m) R_{k-1}(e_3)^{\wedge},\\
    F_{3} =& -\hat{\Omega}_{k-1},\\
    F_{4} =& J^{-1}((J\Omega_{k-1})^\wedge - \hat{\Omega}_{k-1}J).
\end{align*}
To obtain the linearized discrete form of the nonlinear quadrotor model is \( X_k^- = F_{k-1}X_{k-1}^+ + G_{k-1}\mathcal{W}_{k-1}\), where \(\mathcal{W}_{k-1}\in \mathbb{R}^{6}\) and \(\mathcal{V}_{k}\in \mathbb{R}^{6}\) represents the zero-mean Gaussian white process noises.
The matrices \(F_{t_{k-1}}\) and \(G_{t_{k-1}}\) are transformed into their respective discrete forms \(F_{k-1}\) and \(G_{k-1}\) using the following relationship,
\begin{align}
    {F}_{t_{k-1}} &\equiv \left. \frac{\partial{f}}{\partial{X}} \right|_{\dot{X}_{(t)},\mathcal{U}_{(t)}},\quad
    {G}_{t_{k-1}} \equiv \left. \frac{\partial{G_{(t)}}}{\partial{X}} \right|_{\dot{X}_{(t)},\mathcal{U}_{(t)}}, \label{eq:FG_c}\\  
    {F}_{k-1} &= I_{n \times n} + \Delta t {F}_{t_{k-1}}\Psi, \quad {G}_{k-1} = \Delta t\Psi {G}_{t_{k-1}}, \label{eq:FGc1}
\end{align}
where, \(\Psi = I_{n \times n} + \frac{\Delta t}{2}{F}_{t_{k-1}}\), and  \(\Delta t\) is the time step. 
In this study, the continuous-discrete form of the EKF was adapted \cite{Crassidis2004}. Thus, during the flow update stage, the priori estimate of the states \(X^-_k\), is obtained by propagating the previous state estimate \(X^+_{k-1}\), through Equations \eqref{eq:xdot_w1} to \eqref{eq:vdot_w1}, using Runge-Kutta fourth-order integration. Propagation through nonlinear dynamics of the quadrotor, improves the accuracy of the priori estimate. To obtain the priori estimate of the error covariance matrix \(P^-_k\) The previous estimate of the error covariance matrix \(P^+_{k-1}\), is propagated as follows, 
\begin{gather}
     P_k^- = F_{k-1}P_{k-1}F_{k-1}^T + G_{k-1}\mathcal{Q}_kG_{k-1}^T,
    \label{eq:P-kj}
\end{gather}
where, \(\mathcal{Q}_k=\mathbb{E}[\mathcal{W}_k\mathcal{W}_k^T]\in\mathbb{R}^{6 \times 6}\) is the process noise covariance matrix.

During the measurement update stage, the Kalman gain \(K_k \in \mathbb{R}^{15 \times6} \) is calculated as follows,
\begin{align}
    K_k = P_k^-H_k^T[H_k P_k^-H_k^T+\mathcal{R}_k]^{-1},
\end{align}
where, \(H_{k} = \left.\frac{\partial{h}}{\partial{X}} \right|_{\bar{X}_k^-} \in \mathbb{R}^{6 \times 15}\), is the Jacobian of the measurement function, and \(\mathcal{R}_k=\mathbb{E}[\mathcal{V}_k\mathcal{V}_k^T]\in\mathbb{R}^{6 \times 6}\). Linearized form of the general measurement equation can be expressed as follows, 
\begin{align}
Y_k = H_kX_k^- + \mathcal{V}_k,
    \label{eq:flowY}
\end{align}
where, \(\mathcal{V}_{k}\in \mathbb{R}^{6}\) represents the zero-mean Gaussian white measurement noises. 
Full state vector \(X\) is then updated during the measurement update. The attitude \(R^-_k\) and angular velocity priori estimates \(\Omega^-_k\) are updated using angular velocity measurements \(\Omega_{imu}\), while the position \(x^-_k\) and velocity \(v^-_k\) priori estimates are updated using measurements from GPS, \(x_{gps}\). Accelerometer measurement \(a_{imu}\) is only used in updating the wind velocity \(v_{w_k}^-\). The true measurements from the sensors \(Z_k \in \mathbb{R}^6\), are used in updating the posteriori states \(X_k^+\), as follows,
\begin{align}
    \bar{X}_k^+=\bar{X}_k^-+K_k[Z_k-H_k\bar{X}_k^-].
\end{align}
Finally, the posteriori error covariance matrix \(P^+_k\) is updated as follows,
\begin{gather}
    P_k^+=[I-K_kH_k]P^-_k[I-K_kH_k]^T+K_k\mathcal{R}_kK_k^T.
    \label{eq:P+kj}
\end{gather}
Equation \eqref{eq:P+kj}, is Joseph's form for error covariance \cite{Gamagedara2019, Crassidis2004}. This form is used to ensure that the error covariance matrix \(P^+_k\) remains positive definite and symmetric.

\paragraph{UKF}
The UKF is a variation of the Kalman filter designed to address the nonlinearities of a discrete dynamic system. In UKF, a set of strategically chosen points around the mean of the states, known as sigma points, plays a major role in the UKF algorithm. The sigma points are propagated through the nonlinear system functions. Thus, allowing to capture the transformation of the mean and covariance directly. The full state vector \(X=[x, v, \Omega, R, v_w]\in \mathbb{R}^{21}\) is used in the UKF algorithm. As the measurement noise is additive, a non-augmented UKF algorithm was used in this study \cite{Haykin2001}.

During measurement update stage, the posteriori state estimates from the previous time-step are used to generated the sigma points \((\mathcal{X}^i_{k-1} \in \mathbb{R}^{n \times 2n+1}, \quad i=0,...,2n)\), such that \(\mathcal{X}_{k-1} = \bigg[ X_{k-1}^+, \quad X_{k-1}^+ \pm \sqrt{(n + \lambda)P_{k-1}^+} \bigg]\). Here the scaling parameter \(\lambda\in \mathbb{R}\) is calculated such that \(\lambda=\alpha^2(n+\kappa)-n\), where \(\alpha=1\) ensures sufficient spread of the sigma points, and \(\kappa \in \mathbb{R}\) determines the secondary scaling. In this study \(\alpha=1\), to ensure the sigma points are spread sufficiently, \(\kappa =0\) is adopted since additional scaling is unnecessary. A larger \(\lambda\) spreads the points more, captures more nonlinearity, but at a higher risk of the system becoming unstable. There a total of \(n=21\) states, thus a total of \(43\) sigma points are generated. The generated sigma points \(\mathcal{X}^i_{k-1}\), are then propagated through the nonlinear dynamic equations of the quadrotor UAV, Equations \eqref{eq:xdot_w1} to \eqref{eq:vdot_w1}, using Runge-Kutta fourth-order integration. Propagated sigma points are then used to calculated the priori mean estimate of the full state vector \(\bar{X}^-_k \in \mathbb{R}^{21}\), and priori estimate of the error covariance \(P^-_k\) as follows,
\begin{align}
    \bar{X}_k^- &= \sum _{i=0}^{2n} W_i^{mean} \mathcal{X}_k^i, \\
    P_k^- &= \sum_{i=0}^{2n} W_i^{cov}(\mathcal{X}_k^i - \bar{X}_k^-)(\mathcal{X}_k^i - \bar{X}_k^-)^T + \mathcal{Q}_k,
\end{align}
where, \(W_{0}^{mean} = \frac{\lambda}{n + \lambda}\), \(W_{0}^{cov} = \frac{\lambda}{n + \lambda}(1 - \alpha ^2 + \beta)\), and \(W_{i}^{mean} = W_{i}^{cov} = \frac{1}{2(n + \lambda )}\in \mathbb{R}, \quad i = 1, 2, . . ., n\).  The covariance weight for the sigma points are denoted by the parameter \(\beta\) . For Gaussian distributions, \(\beta\) is considered to be 2. 

During the measurement update stage, the priori estimates \(\bar{X}^-_k\) and \(P^-_k\) are updated. For that, first the mean predicted measurement \(\bar{Y}_k^-\in \mathbb{R}^6\) should be obtained. Hence, the sigma points \(\mathcal{X}^i_k\), are transformed using the general measurement equation, \(Y_k = h(X_k, \mathcal{U}_k, \mathcal{V}_k, k)\). The transformed sigma points \(\mathcal{Y}^i_k \in \mathbb{R}^6\) are then used as follows,
\begin{align}
     \bar{Y}_k = \sum_{i=0} ^{2n} W_{i}^{mean} \mathcal{Y}_k^i. 
\end{align}
Subsequently, Kalman gain \(K_k \in \mathbb{R}^{21 \times 6}\) is calculated as follows,
\begin{align}
    P_{yy_k} &= \sum_{i=0}^{2n} W_i^{cov} (\mathcal{Y}_k^i - \bar{Y}_k)(\mathcal{Y}_k^i - \bar{Y}_k)^T +\mathcal{R}_k,\\
    P_{xy_k} &= \sum _{i=0}^{2n} W_i^{cov}(\mathcal{X}_k^i - \bar{X}_k^-)(\mathcal{Y}_k^i - \bar{Y}_k )^T,\\
    K_k &= P_{xy_k} P_{yy_k}^{-1}.
\end{align}
Then using \(\bar{Y}^-_k\) and \(K_k\) the posteriori state estimate \(\bar{X}^+_k\) and the error covariance \(P^+_k\) are calculated as follows,
\begin{align}
    \bar{X}_k^+ = \bar{X}_k^{-} + K_k(Z_k - \bar{Y}_k),\\
    P_k^+ = P_k^- + K_kP_{yy_k}K_k^T.
\end{align}
The true measurements at the current time-step are \(Z_k =[x_{gps}, \Omega_{imu}]\in \mathbb{R}^6\).

\subsection{Case Settings}
The parameters of the quadrotor used in the simulation are included in Table \ref{tab:quadrotor_parameters}. It was assumed that the quadrotor UAV is equipped with an ICM-45686 (6-axis IMU) \cite{icm45686} with gyroscope accuracy \(3.8\;mdps/\sqrt{Hz}\) and accelerometer accuracy \(70\;\mu g/\sqrt{Hz}\), as well as an M10 GPS module \cite{m10gps} with accuracy \(2.0\;CEP\). The corresponding standard deviations are summarized in Table \ref{tab:quadrotor_parameters}. Since the flight tajectory is considered relatively small, the GPS standard deviation was assumed to be \(0.1\;m\).
\begin{table}
\centering
    \caption{Parameters of the quadrotor model}
    \label{tab:quadrotor_parameters}
\begin{tabular}{|p{4.8cm}|p{1.6cm}|c|}
    \hline
    \textbf{Symbol: Description} & \textbf{Value} & \textbf{Unit} \\ \hline
    
    $I_{xx}$: MOI about body frame's x-axis & $0.02$ & $\mathrm{kg{\cdot}m^2}$ \\ \hline
    
    $I_{yy}$: MOI about body frame's y-axis & $0.02$ & $\mathrm{kg{\cdot}m^2}$ \\ \hline
    
    $I_{zz}$: MOI about body frame's z-axis & $0.04$ & $\mathrm{kg{\cdot}m^2}$ \\ \hline
    
    $d$: Distance from rotor arm to the quadrotor's center of gravity & $0.169$ & $m$ \\ \hline
    
    $m$: Quadrotor mass & $2.0$ & $kg$ \\ \hline

    $C_d$: Body drag force coefficient & diag([ 3.265, 3.265, 1.633])$\times10^{-2}$  & - \\ \hline

    $\sigma_{gps}$: GPS standard deviation & $0.1$& $m$\\ \hline

    $\sigma_{imu}$: Gyroscope standard deviation & $0.009$& $ rad/s$\\ \hline

    $\sigma_{acc}$: Accelerometer standard deviation & $0.097$& $ ms^{-2}$\\ \hline

    $g$: Gravitational acceleration & $9.81$ & $ms^{-2}$ \\ \hline
    \end{tabular}
\end{table}
The system was introduced with fixed disturbances \(\Delta_1\) and \(\Delta_2\) as process noise and non-additive, zero-mean, Gaussian noise as the measurement noise, that affects the measurements from the GPS, gyroscope, and accelerometer. Equations \eqref{eq:vdot_w1} and \eqref{eq:Wdot_w1} include the fixed disturbances, \(\Delta _1 = [0.5,  0.8, -1.0], \: \: \Delta _2 = [0.2, 1.0, -0.1]\).

The performance of EKF and UKF estimations was tested using numerical simulations using MATLAB. The quadrotor UAV was tested while hovering, as well as while following a Lissajous trajectory. The performance of the system was observed for \(15\) seconds. at a time-step of \(0.005\) seconds. 
During hovering, the quadrotor started at an initial point \( x_i = [0, 0, 0]^T\) and traveled to  \( x_d = [1, 0, -1]^T\), and hovered at the point. The desired trajectory of the Lissajous curve is \(x_{d(t)} = [ sin(t ), sin(2t),  -1 + 0.2cos(2t)]^T\).
The quadrotor UAV is disturbed with a constant wind of \( v_{wt} = [4, 5, 0]^T\). To gauge the estimation accuracy in a more complex wind field,  a sinusoidal wind field, \(v_{wt} =[5\sin(2\pi f_1 t), 4\sin(2\pi f_2 t), 4\sin(2\pi f_2 t)]^T \), where, \(f_1 = 1/15 \: Hz\) and \(f_2 = 2/15 \: Hz\). 

\section{Results and Discussion}

Wind velocity estimation using quadrotor UAV, is carried out in two stages. First stage is for fixed-disturbance identification. The quadrotor UAV hovers, fly through a straight-line trajectory and through a Lissajous  curve, in the absence of wind disturbances, during which the fixed disturbances are estimated using EKF and UKF. The estimates from each trajectory are then averaged to obtain an final estimate for the fixed disturbances, which are shown in Table \ref{tab:bias_estimate}.
\begin{table}[htb!]
    \centering
    \caption{Estimations for bias of a Quadrotor UAV}
    \label{tab:bias_estimate}
    \begin{tabular}{|c|c|c|c|c|c|c|}
    \hline
    \textbf{Type} & \multicolumn{3}{|c|}{\(\Delta_1\)} & \multicolumn{3}{|c|}{\textbf{\(\Delta_2\)}} \\ \hline
    EKF & 0.516& 0.776& -0.992& 0.200& 1.000& -0.100\\ \hline
    UKF & 0.512& 0.802& -0.978& 0.200& 1.000& -0.100\\ \hline
    True & 0.500& 0.800& -1.000& 0.200& 1.000& -0.100\\ \hline
              \end{tabular}
\end{table}

During the second stage, quadrotor UAV estimates the wind relying on the corresponding estimates from stage one. To assess the performance of quadrotor UAV with EKF and UKF algorithms, the wind velocity estimation is carried out in three different cases.
Wind velocity estimation was carried out in three different cases. 
Case 1: Quadrotor following a Lissajous trajectory under constant wind disturbances, 
Case 2: Quadrotor in hovering under sinusoidal wind disturbances, and 
Case 3: Quadrotor following a Lissajous trajectory under sinusoidal wind disturbances. 
In this work, the drag moment is neglected, considering its negligible effects, compared to the drag force. The measurement noise corrupting the measurements are given in Table \ref{tab:quadrotor_parameters}. 
As listed in Table \ref{tab:quadrotor_parameters}, sensor-based values are used for \(\mathcal{R}_k\) and are adjusted to optimize estimation performance. The process and measurement noise covariances are defined as (\(\mathcal{Q}_{EKF} \in \mathbb{R}^{6 \times 6}\) and \(\mathcal{Q}_{UKF} \in \mathbb{R}^{21 \times 21}\)), (\(\mathcal{R}_{EKF} \in \mathbb{R}^{9 \times 9} \), and \(\mathcal{R}_{UKF} \in \mathbb{R}^{6 \times 6}\)). 
The initial estimates for wind velocity \((v_w)\) contained high-frequency noise. Thus, a second-order Butterworth filter is used as a postprocessing step. The corresponding results, evaluated via RMSE and standard deviation, are presented in Table \ref{tab:Vw_estimate}.

\begin{table}
    \centering
    \caption{3D Wind Velocity Estimation RMSE via EKF and UKF}
    \begin{tabular}{|c|c|c|}
    \hline
         & \textbf{Estimator} & \textbf{Filtered Estimate} \\
         \hline
     \multirow{6}{4em}{\textbf{Case 1}}   & \multirow{3}{4em}{EKF} & \(0.8192 \:(\pm 0.6044)\)\\
                                                &                        & \(0.7955 \:(\pm 0.7416)\)\\
                                                &                        & \(1.3490 \:(\pm 0.8249)\)\\\cline{2-3}
                                                & \multirow{3}{4em}{UKF} & \(0.6238 \:(\pm 0.5370)\) \\
                                                &                        & \(0.8461 \:(\pm 0.7138)\) \\
                                                &                        & \(1.2511 \:(\pm 0.8040)\) \\
    \hline
     \multirow{6}{4em}{\textbf{Case 2}}  & \multirow{3}{4em}{EKF} & \(0.4562 \: (\pm 0.4486)\)\\
                                                &                        & \(0.7322 \: (\pm 0.7271)\)\\
                                                &                        & \(1.1373 \: (\pm 1.0870)\)\\\cline{2-3}
                                                & \multirow{3}{4em}{UKF} & \(0.4525 \: (\pm 0.4505)\) \\
                                                &                        & \(0.6863 \: (\pm 0.6314)\) \\
                                                &                        & \(1.0530 \: (\pm 1.0275)\) \\
    \hline
    \multirow{6}{4em}{\textbf{Case 3}}    &  \multirow{3}{4em}{EKF}& \(0.6935 \:(\pm 0.6857)\)\\
                                                &                        & \(0.6325 \:(\pm 0.6294)\)\\
                                                &                        & \(1.3425 \:(\pm 1.3255)\)\\\cline{2-3}
                                                & \multirow{3}{4em}{UKF} & \(0.6694 \:(\pm 0.6695)\) \\
                                                &                        & \(0.4964 \:(\pm 0.4566)\) \\
                                                &                        & \(1.2463 \:(\pm 1.2442)\) \\
    \hline
    \end{tabular}
    \label{tab:Vw_estimate}
\end{table}
According to the results, both EKF and UKF estimate the wind velocities with highest accuracy in Case 2 (hovering under sinusoidal wind disturbances). Among the three investigated cases, Case 1 (Lissajous trajectory under constant wind disturbances) results in the lowest estimation accuracy for both EKF and UKF. The Case with highest nonlinearities is the Case 3 (Lissajous trajectory under sinusoidal wind disturbances). As the nonlinearity of the trajectory and wind disturbances increases, the UKF tends to consistently outperform the EKF. Overall the accuracy of wind estimation in vertical direction is lower. 

\begin{figure}
    \centering
    \includegraphics[width=1\linewidth]{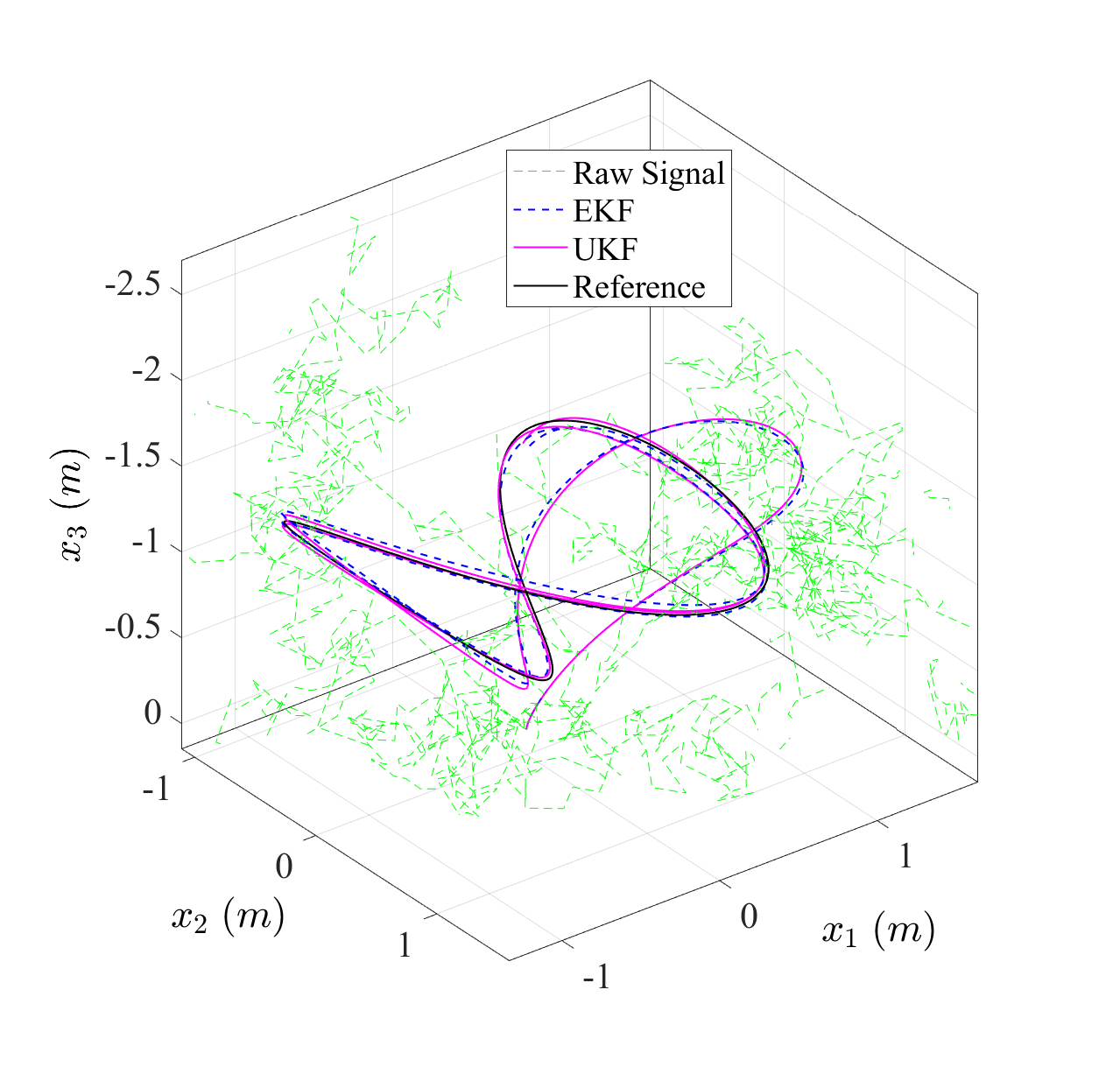}
    \caption{Trajectory of the quadrotor UAV while following a Lissajous trajectory under sinusoidal wind disturbances.}
    \label{fig:IMP3_3D}
\end{figure}

When assessing the trajectory holding capability of the quadrotor UAV under wind disturbances, it was observed that both estimates were capable of following the trajectory, even during the highest nonlinear scenario, Case 3.  Figure \ref{fig:IMP3_3D} shows the trajectory of the quadrotor UAV relying only on sensor measurement, with EKF and with UKF. Both EKF and UKF estimators ensure that the quadrotor UAV follows the intended path with minimum deviation. 

The quadrotor UAV was capable of following the trajectory even during the highest nonlinear scenario, Case 3. Figure \ref{fig:IMP3_3D} shows the trajectory of the quadrotor UAV in 3D space, under sinusoidal wind disturbances. The quadrotor UAV was not capable of adhering to the path when it is relying only on the sensor measurements. Thus, indicating the need for an estimator. Both EKF and UKF estimators ensure that the quadrotor UAV follows the intended path with minimum deviation. 
To evaluate the trajectory holding ability of the two estimators, the RMSE and standard deviations of the position \((x)\), velocity \((v_w)\), angular velocity \((\Omega)\) and attitude \((R)\) were calculated. The results are included in Table \ref{tab:X_estimate}. According to the results, both EKF and UKF estimators are capable of trajectory holding, while the EKF is faster. The orientation error vector was obtained as \(\frac{1}{2}(R^TR_t-(R^TR_t)^T)^\vee\), where \(R\) is the estimated rotation and \(R_t\) is the true rotation.   The symbol \(\vee\) represents the vee map. Errors of the remaining state calculated by computing the difference between and true values. The EKF is capable of performing the estimation sparing only 0.0001 seconds for a single iteration, while UKF used 0.0019 seconds.

\begin{table}[htb!]
    \centering
    \caption{State Estimation RMSE and standard deviation via EKF and UKF for Lissajous Trajectory in a Sinusoidal Wind Field}
    \label{tab:X_estimate}
    \begin{tabular}{|c|c|c|}
    \hline
    \textbf{State}  &  \textbf{EKF} & \textbf{UKF} \\\hline        
    \multirow{3}{4em}{\centering \textbf{\(x\)}\\\((m)\)}  & 0.0172 (\(\pm\) 0.0171)& 0.0115 (\(\pm\) 0.0115)\\ 
                                        & 0.0164 (\(\pm\) 0.0163)& 0.0107 (\(\pm\) 0.0104)\\ 
                                        & 0.0173 (\(\pm\) 0.0174)& 0.0113 (\(\pm\) 0.0113)\\ \hline
    \multirow{3}{4em}{\centering\(v\)\\\((ms^{-1})\)}  & 0.0565 (\(\pm\) 0.0565)& 0.0225 (\(\pm\) 0.0225)\\ 
                                        & 0.0571 (\(\pm\) 0.0571)& 0.0210 (\(\pm\) 0.0210)\\ 
                                        & 0.0322 (\(\pm\) 0.0325)& 0.0261 (\(\pm\) 0.0257)\\ \hline
    \multirow{3}{4em}{\centering\(\Omega\)\\\((rads^{-1})\)}      & 0.0010 (\(\pm\) 0.0003)& 0.0057 (\(\pm\) 0.0057)\\ 
                                                 & 0.0011(\(\pm\) 0.0003)& 0.0057 (\(\pm\) 0.0057)\\ 
                                                 & 0.0009 (\(\pm\) 0.0005)& 0.0057 (\(\pm\) 0.0057)\\ \hline
    \multirow{3}{4em}{\centering\(R\)}      & 0.0003 (\(\pm\) 0.0003)& 0.0010 (\(\pm\) 0.0010)\\ 
                                            & 0.0003 (\(\pm\) 0.0003)& 0.0008 (\(\pm\) 0.0008)\\ 
                                            & 0.0010 (\(\pm\) 0.0009)& 0.0014 (\(\pm\) 0.0011)\\
 \hline
              \end{tabular}
\end{table}

Results from Table \ref{tab:X_estimate} indicate that the UKF is capable of accurately estimating position \((x)\) than the UKF, amid noisy measurements. In addition, the parameter which is totally estimated by the estimator velocity \((v)\), has much better accuracy, when estimated using UKF. However, the EKF shows higher accuracy in estimating the angular velocity \((\Omega)\) and the attitude \((R)\). In terms of computational time, EKF is far more quicker than the UKF.

\section{Conclusion}
This study improves wind velocity estimation and trajectory tracking of quadrotor UAVs using UKF. The estimations relied solely on commonly available onboard sensors. The performance of the UKF was compared to the widely used EKF. The results indicated that the UKF outperformed the EKF as system nonlinearity increased. However, in near-linear scenarios, the EKF achieved similar accuracy with less computation time. In wind velocity estimation, both the UKF and EKF maintained accurate trajectory tracking, despite wind disturbances. The UKF eliminated the need for high-cost sensors by providing cost-effective estimates. It is well-suited for critical tasks like wildfire monitoring, where nonlinearities are significant. For more general tasks, the EKF is computationally faster. Future work will focus on improving the accuracy of vertical wind velocity estimation.

\section*{Acknowledgment}

The authors acknowledge the support of Natural Sciences and Engineering Research Council of Canada (NSERC) and Alberta Innovates.

\bibliographystyle{IEEEtran}
\bibliography{citations}

@article{gonzalez2019sensing,
  title={Sensing wind from quadrotor motion},
  author={Gonz{\'a}lez-Rocha, Javier and Woolsey, Craig A and Sultan, Cornel and De Wekker, Stephan FJ},
  journal={Journal of Guidance, Control, and Dynamics},
  volume={42},
  number={4},
  pages={836--852},
  year={2019},
  publisher={American Institute of Aeronautics and Astronautics}
}

@article{Goodarzi2017,
   author = {Farhad A. Goodarzi and Taeyoung Lee},
   doi = {10.1007/s10846-017-0525-6},
   issn = {15730409},
   issue = {2-4},
   journal = {Journal of Intelligent and Robotic Systems: Theory and Applications},
   month = {12},
   pages = {395-413},
   publisher = {Springer Netherlands},
   title = {Global Formulation of an Extended Kalman Filter on SE(3) for Geometric Control of a Quadrotor UAV},
   volume = {88},
   year = {2017}
}

@inproceedings{Gamagedara2019,
   

   author = {Kanishke Gamagedara and Taeyoung Lee and Murray Snyder},
   doi = {10.2514/6.2019-2377},
   isbn = {9781624105784},
   booktitle = {AIAA Scitech 2019 Forum},
   publisher = {American Institute of Aeronautics and Astronautics Inc, AIAA},
   title = {Real-time kinematics gps based telemetry system for airborne measurements of ship air wake},
   year = {2019}
}

@book{Crassidis2004,
  author        = {Crassidis, J. L. and Junkins, J. L.},
  title         = {Optimal Estimation of Dynamics Systems},
  publisher     = {Chapman \& Hall CRCy},
  year          = {2004},
  Edition        = {Second}
}

@book{Haykin2001,
  
   author = {Simon S.. Haykin},
    title = {Kalman Filtering and Neural Networks},
   publisher = {John Wiley \& Sons},
   year = {2001}
}

@article{Xing2017,
   author = {Zhewen Xing and Yaohong Qu and Youmin Zhang},
   doi = {10.1109/ICAMechS.2017.8316534},
   isbn = {9781538626023},
   issn = {23250690},
   journal = {International Conference on Advanced Mechatronic Systems, ICAMechS},
   pages = {196-201},
   title = {Shear wind estimation with quadrotor UAVs using Kalman filtering regressing method},
   volume = {2017-Decem},
   year = {2017}
}

@article{Xiang2016,
   author = {Xingyu Xiang and Zhonghai Wang and Zijian Mo and Genshe Chen and Khanh Pham and Erik Blasch},
   doi = {10.1109/DASC.2016.7778071},
   isbn = {9781509056002},
   issn = {21557209},
   journal = {AIAA/IEEE Digital Avionics Systems Conference - Proceedings},
   pages = {1-6},
   publisher = {IEEE},
   title = {Wind field estimation through autonomous quadcopter avionics},
   volume = {2016-Decem},
   year = {2016}
}

@article{Chen2022,
   author = {Hao Chen and He Bai and Clark N. Taylor},
   doi = {10.23919/ACC53348.2022.9867417},
   isbn = {9781665451963},
   issn = {07431619},
   journal = {Proceedings of the American Control Conference},
   pages = {1236-1241},
   publisher = {American Automatic Control Council},
   title = {Invariant-EKF design for quadcopter wind estimation},
   volume = {2022-June},
   year = {2022}
}

@article{Shastry2023,
   author = {Animesh K. Shastry and Derek A. Paley},
   doi = {10.1002/rnc.6935},
   issn = {10991239},
   issue = {17},
   journal = {International Journal of Robust and Nonlinear Control},
   month = {11},
   pages = {10451-10467},
   publisher = {John Wiley and Sons Ltd},
   title = {System identification for high-performance UAV control in wind},
   volume = {33},
   year = {2023}
}

@inproceedings{Rhudy2019,
   author = {Matthew Rhudy and Jason Gross and Yu Gu},
   doi = {10.2514/6.2019-3111},
   isbn = {9781624105890},
   booktitle = {AIAA Aviation 2019 Forum},
   pages = {1-9},
   publisher = {American Institute of Aeronautics and Astronautics Inc, AIAA},
   title = {Stochastic wind modeling and estimation for unmanned aircraft systems},
   year = {2019}
}

@inproceedings{Rhudy2014,
   author = {Matthew B. Rhudy and Yu Gu and Haiyang Chao},
   city = {Atlanta, GA},
   doi = {10.2514/6.2014-2647},
   booktitle = {AIAA Modeling and Simulation Technologies Conference},
pages = {1-9},
   month = {6},
   publisher = {American Institute of Aeronautics and Astronautics (AIAA)},
   title = {Wind Field Velocity and Acceleration Estimation Using a Small UAV},
   year = {2014}
}

@inproceedings{Condomines2015,
   author = {Jean-Philippe Condomines and Murat Bronz and Gautier Hattenberger and Jean-François Erdelyi},
   city = {Aachen, Germany},
   booktitle = {IMAV 2015: International Micro Air Vehicles Conference and Flight Competition},
   month = {9},
   pages = {},
   publisher = {HAL open science},
   title = {Experimental Wind Field Estimation and Aircraft Identification},
   url = {http://www.researchgate.net/publication/282003424},
   year = {2015}
}

@article{Rhudy2017,
   author = {Matthew B. Rhudy and Yu Gu and Jason N. Gross and Haiyang Chao},
   doi = {10.1109/TAES.2017.2649218},
   issn = {00189251},
   issue = {1},
   journal = {IEEE Transactions on Aerospace and Electronic Systems},
   month = {2},
   pages = {55-66},
   publisher = {Institute of Electrical and Electronics Engineers Inc.},
   title = {Onboard Wind Velocity Estimation Comparison for Unmanned Aircraft Systems},
   volume = {53},
   year = {2017}
}

@inproceedings{LNCSikkel2016,
   author = {L.N.C. Sikkel and G.C.H.E. de Croon and C. De Wagter and Q.P. Chu},
   city = {Daejeon, Korea},
   isbn = {9781509037629},
   booktitle = {2016 IEEE/RSJ International Conference on Intelligent Robots and Systems},
   month = {10},
   pages = {2141-2146},
   publisher = {IEEE},
   title = {IROS 2016 : 2016 IEEE/RSJ International Conference on Intelligent Robots and Systems : October 9-14, 2016, Daejeon Convention Center, Daejeon, Korea},
   year = {2016}
}

@misc{icm45686,
  title        = {{ICM-45686}},
  author       = {{TDK InvenSense}},
  howpublished = {\url{https://invensense.tdk.com}},
  note         = {Accessed: 2025-06-25},
  year         = {2025}
}

@misc{m10gps,
  title        = {{M10 GPS}},
  author       = {{Holybro Store}},
  howpublished = {\url{https://holybro.com}},
  note         = {Accessed: 2025-06-25},
  year         = {2025}
}

@article{Neumann2015,
   author = {Patrick P. Neumann and Matthias Bartholmai},
   doi = {10.1016/j.sna.2015.09.036},
   issn = {09244247},
   journal = {Sensors and Actuators, A: Physical},
   month = {11},
   pages = {300-310},
   publisher = {Elsevier},
   title = {Real-time wind estimation on a micro unmanned aerial vehicle using its inertial measurement unit},
   volume = {235},
   year = {2015}
}

@inproceedings{Gonzlez-Rocha2019,
   author = {Javier González-Rocha and Craig A. Woolsey and Cornel Sultan and Stephan F.J. De Wekker},
   city = {San Diego, California},
   doi = {10.2514/6.2019-1598},
   isbn = {9781624105784},
   booktitle = {AIAA Scitech 2019 Forum},
   month = {1},
   publisher = {American Institute of Aeronautics and Astronautics Inc, AIAA},
   title = {Model-based wind profiling in the lower atmosphere with multirotor UAS},
   year = {2019}
}

@article{Shastry2021,
   author = {Animesh Shastry and Derek A. Paley},
   doi = {10.1109/LCSYS.2020.3044491},
   issn = {24751456},
   issue = {5},
   journal = {IEEE Control Systems Letters},
   month = {11},
   pages = {1801-1806},
   publisher = {Institute of Electrical and Electronics Engineers Inc.},
   title = {UAV State and Parameter Estimation in Wind Using Calibration Trajectories Optimized for Observability},
   volume = {5},
   year = {2021}
}

\end{document}